\documentclass[11pt,letterpaper]{article}
\usepackage{naaclhlt2016}
\usepackage{times}
\usepackage{latexsym}
\usepackage{amssymb}

\naaclfinalcopy % Uncomment this line for the final submission
\def\naaclpaperid{***} %  Enter the naacl Paper ID here

\title{INSIGHT-1 at SemEval-2016 Task 4: Convolutional Neural Networks for Sentiment Classification and Quantification}

\author{Sebastian Ruder\textsuperscript{1}\textsuperscript{2} \\
	  \And
	Parsa Ghaffari\textsuperscript{2}\\
	\And
	John G. Breslin\textsuperscript{1}
	\AND
	    \normalfont{\textsuperscript{1}Insight Centre for Data Analytics}\\
	    National University of Ireland, Galway\\
	    {\tt firstname.lastname@insight-centre.org}
	    \AND
	    \normalfont{\textsuperscript{2}Aylien Ltd.}\\
	    Dublin, Ireland\\
	    {\tt firstname@aylien.com}
  }

\date{}

\begin{document}

\maketitle

\begin{abstract}
This paper describes our deep learning-based approach to sentiment analysis in Twitter as part of SemEval-2016 Task 4. We use a convolutional neural network to determine sentiment and participate in all subtasks, i.e. two-point, three-point, and five-point scale sentiment classification and two-point and five-point scale sentiment quantification. We achieve competitive results for two-point scale sentiment classification and quantification, ranking fifth and a close fourth (third and second by alternative metrics) respectively despite using only pre-trained embeddings that contain no sentiment information. We achieve good performance on three-point scale sentiment classification, ranking eighth out of 35, while performing poorly on five-point scale sentiment classification and quantification. An error analysis reveals that this is due to low expressiveness of the model to capture negative sentiment as well as an inability to take into account ordinal information. We propose improvements in order to address these and other issues.
\end{abstract}

\section{Introduction}

Social media allows hundreds of millions of people to interact and engage with each other, while expressing their thoughts about the things that move them. Sentiment analysis \cite{Pang2008} allows us to gain insights about opinions towards persons, objects, and events in the public eye and is used nowadays to gauge public opinion towards companies or products, to analyze customer satisfaction, and to detect trends.

Its immediacy allowed Twitter to become an important platform for expressing opinions and public discourse, while the accessibility of large quantities of data in turn made it the focal point of social media sentiment analysis research.

Recently, deep learning-based approaches have demonstrated remarkable results for text classification and sentiment analysis \cite{Kim2014} and have performed well for phrase-level and message-level sentiment classification \cite{Severyn2015a}.

Past SemEval competitions in Twitter sentiment analysis \cite{Rosenthal2014,Rosenthal2015} have contributed to shape research in this field. SemEval-2016 Task 4 \cite{SemEval:2016:task4} is no exception, as it introduces both quantification and five-point-scale classification tasks, neither of which have been tackled with deep learning-based approaches before.

We apply our deep learning-based model for sentiment analysis to all subtasks of SemEval-2016 Task 4: three-point scale message polarity classification (subtask A), two-point and five-point scale topic sentiment classification (subtasks B and C respectively), and two-point and five-point scale topic sentiment quantification (subtasks D and E respectively).

Our model achieves excellent results for subtasks B and D, ranks competitively for subtask A, while performing poorly for subtasks C and E. We perform an error analysis of our model to obtain a better understanding of strengths and weaknesses of a deep learning-based approach particularly for these new tasks and subsequently propose improvements.

\section{Related work}

Deep-learning based approaches have recently dominated the state-of-the-art in sentiment analysis. Kim \shortcite{Kim2014} uses a one-layer convolutional neural network to achieve top performance on various sentiment analysis datasets, demonstrating the utility of pre-trained embeddings.

State-of-the-art models in Twitter sentiment analysis leverage large amounts of data accessible on Twitter to further enhance their embeddings by treating smileys as noisy labels \cite{Go2009}: Tang et al. \shortcite{Tang2014a} learn sentiment-specific word embeddings from such distantly supervised data and use these as features for supervised classification, while Severyn and Moschitti \shortcite{Severyn2015a} use distantly supervised data to fine-tune the embeddings of a convolutional neural network.

In contrast, we observe distantly supervised data not to be as important for some tasks as long as sufficient training data is available.

\section{Model}

The model architecture we use is an extension of the CNN structure used by Collobert et al. \shortcite{Collobert2011a}.

The model takes as input a text, which is padded to length $n$. We represent the text as a concatentation of its word embeddings $x_{1:n}$ where $x_i \in \mathbb{R}^k$ is the $k$-dimensional vector of the $i$-th word in the text.

The convolutional layer slides filters of different window sizes over the word embeddings. Each filter with weights $w \in \mathbb{R}^{hk}$ generates a new feature $c_i$ for a window of $h$ words according to the following operation:

\begin{equation} \label{eq:featuremap}
c_i = f(w \cdot x_{i:i+h-1} + b) 
\end{equation}

Note that $b \in \mathbb{R}$ is a bias term and $f$ is a non-linear function, ReLU \cite{Nair2010} in our case. The application of the filter over each possible window of $h$ words or characters in the sentence produces the following feature map:

\begin{equation} 
c = [c_1, c_2, ..., c_{n-h+1}]
\end{equation}

Max-over-time pooling in turn condenses this feature vector to its most important feature by taking its maximum value and naturally deals with variable input lengths.

A final softmax layer takes the concatenation of the maximum values of the feature maps produced by all filters and outputs a probability distribution over all output classes.

\section{Methodology}

\subsection{Datasets}

For every subtask, the organizers provide a training, development, and development test set for training and tuning. We use the concatentation of the training and development test set for each subtask for training and use the development set for validation. 

Additionally, the organizers make training and development data from SemEval-2013 and trial data from 2016 available that can be used for training and tuning for subtask A and subtasks B, C, D, and E respectively.
We experiment with adding these datasets to the respective subtask. Interestingly, adding them slightly increases loss on the validation set, while providing a significant performance boost on past development test sets, which we view as a proxy for performance on the 2016 test set. For this reason, we include these datasets for training of all our models.

We notably do not select the model that achieves the lowest loss on the validation set, but choose the one that maximizes the $F_1^{PN}$ score, i.e. the arithmetic mean of the $F_1$ of positive and negative tweets, which has historically been used to evaluate the SemEval message polarity classification subtask. We observe that the lowest loss does not necessarily lead to the lowest $F_1^{PN}$, as it does not include $F_1$ of neutral tweets.

\subsection{Pre-processing}

For pre-processing, we use a script adapted from the pre-processing script\footnote{\texttt{http://nlp.stanford.edu/projects/glove/
preprocess-twitter.rb}} used for training GloVe vectors \cite{Pennington2014}. Besides normalizing urls and mentions, we notably normalize happy and sad smileys, extract hashtags, and insert tags for repeated, elongated, and all caps characters.

\subsection{Word embeddings}

Past research \cite{Kim2014,Severyn2015a} found a good initialization of word embeddings to be crucial in training an accurate sentiment model.

We thus evaluate the following evaluation schemes: random initialization, initialization using pre-trained GloVe vectors, fine-tuning pre-trained embeddings on a distantly supervised corpus \cite{Severyn2015a}, and fine-tuning pre-trained embeddings on 40k tweets with crowd-sourced Twitter annotations. Perhaps counter-intuitively, we find that fine-tuning embeddings on a distantly supervised or crowd-sourced corpus does not improve performance on past development test sets when including the additionally provided data for training. We hypothesize that additional training data facilitates learning of the underlying semantics, thereby reducing the need for sentiment-specific embeddings. Our scores partially echo this theory.

For this reason, we initialize our word embeddings simply with 200-dimensional GloVe vectors trained on 2B tweets. Word embeddings for unknown words are initialized randomly.

\subsection{Hyperparameters and pre-processing}

We tune hyperparameters over a wide range of values via random search on the validation set. We find that the following hyperparameters, which are similar to ones used by Kim \shortcite{Kim2014}, yield the best performance across all subtasks: mini-batch size of 10, maximum sentence length of 50 tokens, word embedding size of 200 dimensions, dropout rate of 0.3, $l_2$ regularization of 0.01, filter lengths of 3, 4, and 5 with 100 filter maps each.

We train for 15 epochs using mini-batch stochastic gradient descent, the Adadelta update rule \cite{Zeiler2012}, and early stopping.

\subsection{Task adaptation and quantification}

To adapt our model to the different tasks, we simply adjust the number of output neurons to conform to the scale used in the task at hand (two-point scale in subtasks B and D, three-point scale in subtask A, five-point scale in subtasks C and E).

We perform a simple quantification for subtasks D and E by aggregating the classified tweets for each topic and reporting their distribution across sentiments. We would thus expect our results on subtasks B and D and results on subtasks C and E to be closely correlated.

\section{Evaluation}

We report results of our model in Tables \ref{tab:results_a} and \ref{tab:progress_a} (subtask A), Table \ref{tab:results_b} (subtask B), Tables \ref{tab:results_c} and \ref{tab:mae_results_c} (subtask C), Table \ref{tab:results_d} (subtask D), and Table \ref{tab:results_e} (subtask E). For some subtasks, the organizers make available alternative metrics. We observe that the choice of the scoring metric influences results considerably, with our system always placing higher if ranked by one of the alternative metrics.

\textbf{Subtask A.} We obtain competitive performance on subtask A in Table \ref{tab:results_a}. Analysis of results on the progress test sets in Table \ref{tab:progress_a} reveals that our system achieves competitive $F_1$ scores for positive and neutral tweets, but only low $F_1$ scores for negative tweets due to low recall. This is mirrored in Table \ref{tab:results_a}, where we rank higher for accuracy than for recall. The scoring metric for subtask A, $F_1^{POS}$ accentuates $F_1$ for positive and negative tweets, thereby ignoring our good performance on neutral tweets and leading to only mediocre ranks on the progress test sets for our system.

\begin{table}[]
\centering
\begin{tabular}{c | c | c | c}
\textbf{Metric} & \textbf{Our score} & \textbf{Best score} & \textbf{Rank} \\\hline
$F_1^{PN}$ & \textbf{0.593} & \textbf{0.633} & \textbf{8/34} \\
$R^{PN}$ & 0.616 & 0.670 & 12/34 \\
$Acc^{PN}$ & 0.635 & 0.646 & 5/34
\end{tabular}
\caption{Our score and rank for subtask A for each metric compared to the best team's score (results for official metric in bold).}
\label{tab:results_a}
\end{table}

\begin{table}[]
\centering
\begin{tabular}{c | c c | c c c | c}
 & \multicolumn{2}{c}{\textbf{2013}} & \multicolumn{3}{| c |}{\textbf{2014}} & \textbf{2015} \\
 & \textbf{TW} & \textbf{SMS} & \textbf{TW} & \textbf{TW} & \textbf{LJ} & \textbf{TW} \\
 &  &  &  & \textbf{/s} &  &  \\\hline
+ & 72.49 & 66.73 & 76.84 & 64.52 & 69.08 & 65.56 \\
- & 47.97 & 49.65 & 51.95 & 13.64 & 50.00 & 53.00 \\
= & 67.53 & 77.83 & 65.51 & 45.71 & 67.28 & 65.23
\end{tabular}
\caption{$F_1$ scores of our model for positive, negative, and neutral tweets for each progress dataset of subtask A. TW: Tweet. /s: sarcasm. LJ: Live Journal. +: positive. -: negative. =: neutral.}
\label{tab:progress_a}
\end{table}

\textbf{Subtasks B and D.} We achieve a competitive fifth rank for subtask B by the official recall metric in Table \ref{tab:results_b}. However, ranked by $F_1$ (as in subtask A), we place third -- and second if ranked by accuracy. Similarly, for subtask D, we rank fourth (with a differential of 0.001 to the second rank) by $KLD$, but second and first if ranked by $AE$ and $RAE$ respectively. Jointly, these results demonstrate that classification performance is a good indicator for quantification without using any more sophisticated quantification methods. These results are in line with past research \cite{Kim2014} showcasing that even a conceptually simple neural network-based approach can achieve excellent results given enough training data per class. These results also highlight that embeddings trained using distant supervision, which should be particularly helpful for this task as they are fine-tuned using the same classes, i.e. positive and negative, are not necessary given enough data.

\begin{table}[]
\centering
\begin{tabular}{c | c | c | c}
\textbf{Metric} & \textbf{Our score} & \textbf{Best score} & \textbf{Rank} \\\hline
$R$ & \textbf{0.767} & \textbf{0.797} & \textbf{5/19} \\
$F_1$ & 0.786 & 0.799 & 3/19 \\
$Acc$ & 0.864 & 0.870 & 2/19
\end{tabular}
\caption{Our score and rank for subtask B for each metric compared to the best team's score (results for official metric in bold).}
\label{tab:results_b}
\end{table}

\begin{table}[]
\centering
\begin{tabular}{c | c | c | c}
\textbf{Metric} & \textbf{Our score} & \textbf{Best score} & \textbf{Rank} \\\hline
$KLD$ & \textbf{0.054} & \textbf{0.034} & \textbf{4/14} \\
$AE$ & 0.085 & 0.074 & 2/14 \\
$RAE$ & 0.423 & 0.423 & 1/14
\end{tabular}
\caption{Our score and rank for subtask D for each metric compared to the best team's score (results for official metric in bold).}
\label{tab:results_d}
\end{table}

\textbf{Subtasks C and E.} We achieve mediocre results for subtask C in Table \ref{tab:results_c}, only ranking sixth -- however, placing third by the alternative metric. Similarly, we only achieve an unsatisfactory eighth rank for subtask E in Table \ref{tab:results_e}. An error analysis for subtask C in Table \ref{tab:mae_results_c} reveals that the model is able to differentiate between neutral, positive, and very positive tweets with good accuracy. However, similarly to results in subtask A, we find that it lacks expressiveness for negative sentiment and completely fails to capture very negative tweets due to their low number in the training data. Additionally, it is unable to take into account sentiment order to reduce error for very positive and very negative tweets.

\begin{table}[]
\centering
\begin{tabular}{c | c | c | c}
\textbf{Metric} & \textbf{Our score} & \textbf{Best score} & \textbf{Rank} \\\hline
$MAE^M$ & \textbf{1.006} & \textbf{0.719} & \textbf{6/11} \\
$MAE^\mu$ & 0.607 & 0.580 & 3/11
\end{tabular}
\caption{Our score and rank for subtask C for each metric compared to the best team's score (results for official metric in bold).}
\label{tab:results_c}
\end{table}

\begin{table}[]
\centering
\begin{tabular}{l | c | c | c | c | c }
Sentiment & -2 & -1 & 0 & 1 & 2 \\\hline
$MAE^M$ & 2.09 & 1.29 & 0.78 & 0.17 & 0.71
\end{tabular}
\caption{Macro-averaged mean absolute error ($MAE^M$) of our model for each sentiment class for subtask C. Lower error is better.}
\label{tab:mae_results_c}
\end{table}

\begin{table}[]
\centering
\begin{tabular}{c | c | c | c}
\textbf{Metric} & \textbf{Our score} & \textbf{Best score} & \textbf{Rank} \\\hline
$EMD$ & \textbf{0.366} & \textbf{0.243} & \textbf{8/10}
\end{tabular}
\caption{Our score and rank for subtask E compared to the best team's score.}
\label{tab:results_e}
\end{table}

\subsection{Improvements}

We propose different improvements to enable the model to better deal with some of the encountered challenges.

\textbf{Negative sentiment.} The easiest way to enable our model to better capture negative sentiment is to include more negative tweets in the training data. Additionally, using distantly supervised data for fine-tuning embeddings would likely have helped to mitigate this deficit. In order to allow the model to better differentiate between different sentiments on a five-point scale, it would be interesting to evaluate ways to create a more fine-grained distantly supervised corpus using e.g. a wider range of smileys and emoticons or certain hashtags indicating a high degree of elation or distress.

\textbf{Ordinal classification.} Instead of treating all classes as independent, we can enable the model to take into account ordinal information by simply modifying the labels as in \cite{Cheng2008}. A more sophisticated approach would organically integrate label-dependence into the network.

\textbf{Quantification.} Instead of deriving the topic-level sentiment distribution by predicting tweet-level sentiment, we can directly minimize the Kullback-Leibler divergence for each topic. If the feedback from optimizing this objective proves to be too indirect to provide sufficient signals, we can jointly optimize tweet-level as well as topic-level sentiment as in \cite{Kotzias2015}.

\section{Conclusion}

In this paper, we have presented our deep learning-based approach to Twitter sentiment analysis for two-point, three-point, and five-point scale sentiment classification and two-point and five-point scale sentiment quantification. We reviewed the different aspects we took into consideration in creating our model. We rank fifth and a close fourth (third and second by alternative metrics) on two-point scale classification and quantification despite using only pre-trained embeddings that contain no sentiment information. We analysed our weaker performance on three-point scale sentiment classification and five-point scale sentiment classification and quantification and found that the model lacks expressiveness to capture negative sentiment and is unable to take into account class order. Finally, we proposed improvements to resolve these deficits.

\section*{Acknowledgments}

This project has emanated from research conducted with the financial support of the Irish Research Council (IRC) under Grant Number EBPPG/2014/30 and with Aylien Ltd. as Enterprise Partner. This publication has emanated from research supported in part by a research grant from Science Foundation Ireland (SFI) under Grant Number SFI/12/RC/2289.

\bibliography{semeval_2016_task_4}
\bibliographystyle{naaclhlt2016}

\end{document}